# Wind Speed Prediction and Visualization Using Long Short-Term Memory Networks (LSTM)

Md Amimul Ehsan
Dept. of Electrical and Computer Eng.
Univ. of the District of Columbia
Washington, D.C. 20008 USA
mdamimul.ehsan@udc.edu

Amir Shahirinia
Dept. of Electrical and Computer Eng.
Univ. of the District of Columbia
Washington, D.C. 20008 USA
amir.shahirinia@udc.edu

Nian Zhang
Dept. of Electrical and Computer Eng.
Univ. of the District of Columbia
Washington, D.C. 20008 USA
nzhang@udc.edu

Timothy Oladunni
Dept. of Computer Science &
Information technology
Univ. of the District of Columbia
timothy.oladunni@udc.edu

*Abstract*— Climate change is one of the most concerning issues of this century. Emission from electric power generation is a crucial factor that drives the concern to the next level. Renewable energy sources are widespread and available globally, however, one of the major challenges is to understand their characteristics in a more informative way. This paper proposes the prediction of wind speed that simplifies wind farm planning and feasibility study. Twelve artificial intelligence algorithms were used for wind speed prediction from collected meteorological parameters. The model performances were compared to determine the wind speed prediction accuracy. The results show a deep learning approach, long short-term memory (LSTM) outperforms other models with the highest accuracy of 97.8%.

## I. INTRODUCTION

Wind speed prediction is a regression problem where predictors, in this case, are the meteorological parameters, and the response variable is the wind speed at 80m height. In general, regression is a classical problem both in statistics and machine learning. Usually, statistical methods are to find the inference while machine learning makes the prediction [1]. Then again, they intersect in some cases and do serve a similar purpose, for example- linear regression. However, statistical learning relies on distributions, while machine learning is an empirical process and requires data [2]. The statistical approach, thus, considers how data is collected or generated while machine learning may result in accurate prediction without knowing much about the underlying aspects of data. Another line of the boundary is the shape or volume of data. While the statistical approach is very robust about the number of samples (as it considers the distribution of the data), machine learning is more applicable when the dataset is wide [2]. However, sometimes they are used interchangeably, and statistics are the backbone of machine learning [3]. Besides, some machine learning algorithms use the same bootstrapping methods as statistical models. Besides, researchers are using deep learning for similar prediction problems [4]. Artificial Neural network (ANN) based models usually yield benefits in prediction tasks than statistical models due to its robustness towards the nature of data, especially when there are missing values, or the dataset is not well preprocessed, raw, and large data [5]. Thus, many machine learning regression algorithms use statistical techniques in innovative ways, while deep learning neural network approaches are efficient for analogous tasks. The reason, however, behind the growing popularity of machine learning or deep learning (artificial intelligence, in general) is the availability of computational resources [6]. Therefore, the larger dataset is not a critical issue to work with, which was challenging in the past. In this research, we have considered both approaches, machine learning, and deep learning for our wind speed prediction problem.

The placement of a wind turbine for wind power generation is often a challenging step due to the varying nature of wind speed from a location/height to another location/height [7]. Measuring wind speed at the level of turbine hub height is both expensive and requires continuous maintenance. Meteorological parameters also play a vital role in the wind characteristics. Therefore, wind speed profiling with the variation of meteorological parameters has been a research problem that leads to the prediction of wind speed of a certain location based on those parameters [8-9]. Therefore, utilization of easy to access parameters in a low elevation to predict corresponding wind speed at a higher height is a practical approach. Literature show there have been statistical approaches for wind speed application, while artificial intelligence- deep learning and machine learning are being considered recently [10-11]. In addition to machine learning regression algorithms, neural network-based deep learning techniques are getting attention for alike problems due to higher accuracy.

Deep Neural Network (DNN) can map features from raw data to provide regularization, thus minimizes the variance in each layer [12]. This capability makes DNN suitable for prediction problems. The prediction accuracy is greatly dependent on efficient feature extraction of time series data. Literature [13] shows- Convolutional Neural Network (CNN) creates filters to automate such function that makes it widely applicable for prediction. However, it can represent short term dependence while wind speed comprises both short and long-term dependence fundamentally [14]. Therefore, long short-term memory (LSTM) seems more effective for wind speed prediction. LSTM is a form of Recurrent Neural Network (RNN) that is capable of learning long-term dependencies to make a prediction [15].

The rest of this paper is organized as follows. In Section 2, deep learning algorithms including deep neural networks (DNN), convolutional neural network (CNN), recurrent neural network (RNN) in the form of long short-term memory (LSTM) are introduced. In Section 3, the performance evaluation is presented. In Section 4, The National Renewable Energy Laboratory (NREL) data set is described. The simulations and experimental results are demonstrated. In Section 5, the conclusions are given.

## II. DEEP LEARNING APPROACH

Deep learning algorithms are now applied to solve problems of a diverse nature, including prediction [16]. Therefore, we are considering deep learning algorithms for this research. Firstly, we would like to review a few basics of deep learning. The building blocks of deep learning or artificial neural networks are called perceptron, which mimics an equivalent functionality (in computation) as neuron (a biological cell of the nervous system that uniquely communicates with each other) [17].

Now, perceptrons or artificial neurons receive input signals $(x_1, x_2, \ldots, x_m)$, multiply each input by weight $(w_1, w_2, \ldots, w_m)$, add them together with a pre-determined bias, and pass through the activation function, $f(x)$. The signal goes to output as 0 or 1 based on the activation function threshold value. A perceptron with inputs, weights, summation and bias, activation function, and output all together forms a single layer perceptron [18]. However, in common neural network diagrams, only input and output layers are shown. In a practical neural network, hidden layers are added between the input and output layers. The number of hidden layers is a hyperparameter and usually determined by evaluating the model performance. If the neural network has a single hidden layer, the model is called a shallow neural network, while a deep neural network (DNN) consists of several hidden layers [17]. In this research, we have considered DNN, convolutional neural network (CNN), and recurrent neural network (RNN)- in the form of long short-term memory (LSTM), all of which will be discussed in the following sections.

### A. Deep Neural Network (DNN)

DNN is composed of three neural network layers, namely- an input layer, hidden layer(s), and an output layer. The (number of hidden layers) is tuned through trial and error [17]. Figure 1 illustrates such a model structure with two hidden layers consisting of three neurons each, five input neurons, and one output neuron. The number of neurons depends on the number of inputs and outputs. In Figure 1,
Inputs: [$x_1, x_2, x_3, x_4, x_5$]
Hidden layer weights: $h$
Output: $\hat{y}$

A simplified DNN kernel is formulated in (1) that considers linear modeling. $x$, $W$, and $c$ symbolize input, weights, and bias, respectively, while $w$ and $b$ are linear model parameters. The hidden layer parameter $h$ is shown in (2), where $g$ is the activation function. For DNN modeling, ReLu (3) is used as the hidden layer activation function.

$$f(x;W,c,w,b) = w^T \max\{0, W^T + c\} + b \qquad (1)$$
$$h = g(W^T x + c) \qquad (2)$$
$$f(x) = \max(0, x) \qquad (3)$$

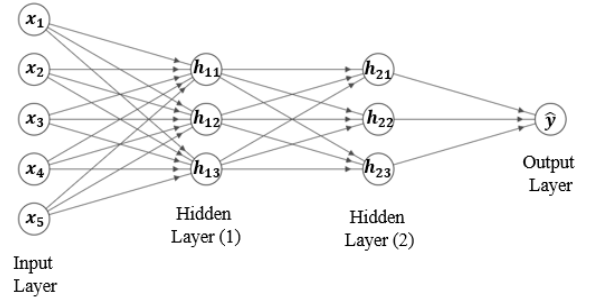

Figure 1 Simplified Architecture of a Deep Neural Network

### B. Convolutional Neural Network (CNN)

CNN, also known as ConvNet, is one way to solve the issue with DNN using convolution rather than matrix multiplication [19]. In other words, CNN is the regularized version of DNN to ensure model robustness towards overfitting. CNN is very popular for image processing; however, in the prediction problem, it is also utilized [20]. In this research, we are using 1D CNN for wind speed prediction. The characteristics and approaches are the same for all CNNs, regardless of dimensionality [21]. The architecture of CNN (Figure 2 shows for 1D CNN) consists of a convolution layer, pooling layer, and a fully connected neural network layer, thus, incorporates local receptive fields to ponder the spatial information, shared weights, and pooling to consider the summary statistics in the output.

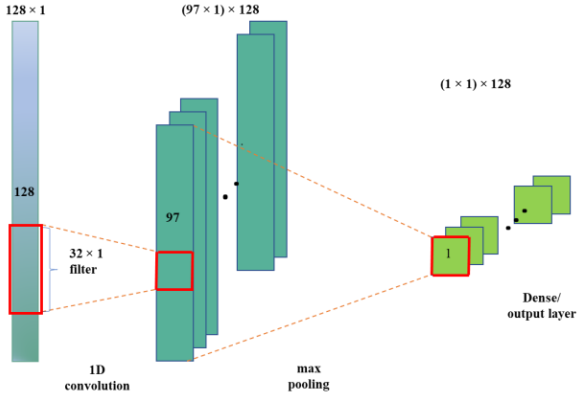

Figure 2 The Architecture of 1D Convolution Neural Network

*C. Recurrent Neural Network (RNN) - LSTM*

Long Short-Term Memory Networks (LSTM), a form of gated RNN, is proposed to implement. LSTM introduces self-loops to produce paths where the gradient can flow for a long duration; thus, it is capable of learning long-term dependencies [17]. LSTMs are explicitly designed to avoid the long-term dependency problem, as illustrated in Figure 3. The equations describing the operations are listed below.

$$f(t) = \sigma_g(W_f x_t + U_f h_{t-1} + b_f) \quad (4)$$

$$i_t = \sigma_g(W_i x_t + U_i h_{t-1} + b_i) \quad (5)$$

$$o_t = \sigma_g(W_o x_t + U_o h_{t-1} + b_o) \quad (6)$$

$$c_t = f_t \circ c_{t-1} + i_t \circ \sigma_c(W_c x_t + U_c h_{t-1} + b_c) \quad (7)$$

$$h_t = o_t \circ \sigma_h(c_t) \quad (8)$$

Where,

$x_t \in \Re^d$ : Input vector to the LSTM unit

$f_t \in \Re^h$ : Forget states activation vector

$i_t \in \Re^h$ : Input/update gate's activation vector

$o_t \in \Re^h$ : Output gate's activation vector

$h_t \in \Re^h$ : Hidden state vector

$c_t \in \Re^h$ : Cell state vector

$W \in \Re^{hxd}, U \in \Re^{hxh}, b \in \Re^h$ : Weight matrices and bias vector parameters which need to be learned during the training

$\sigma_g$ : Sigmoid function

$\sigma_c$, $\sigma_g$ : hyperbolic tangent function

### III. PERFORMANCE EVALUATION

Some commonly used accuracy parameters are employed to evaluate how well a model is performing to predict the intended parameter [22]. Mean absolute error (*MAE*), mean square error (*MSE)*, median absolute error (*MedAE)*, and R-square *(R2)* scores are considered to investigate the model performances on the test set.

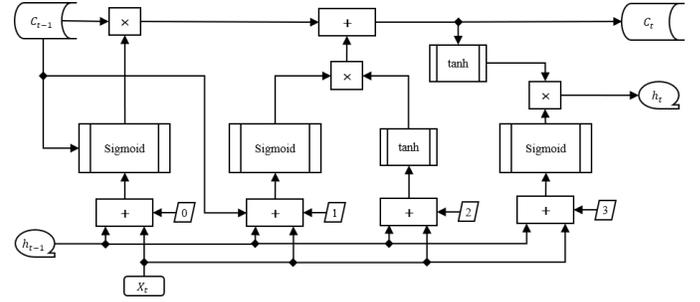

Figure 3 Block Diagram of LSTM Operations

*MAE* is the average of the absolute values of the error (the difference between actual response ( $y_i$ ) and predicted response ( $\hat{y}_l$ ). As described by (9), n is the number of total input sets. The lower this value is, the better the model performance, while the desired is 0.

$$MAE = \frac{\sum_{i=1}^{n}|y_i - \hat{y}_l|}{n} \quad (9)$$

*MSE* is the mean of the square of error terms. Similarly, to *MAE*, it is desired to have 0 or close value for this term. The formula for this measure is in (10).

$$MSE = \frac{\sum_{i=1}^{n}(y_i - \hat{y}_l)^2}{n} \quad (10)$$

*MedAE* is the median of all the error terms, defined in (11), thus effective to deal with outliers' effect in the model performance.

$$MedAE = median(|y_1 - \hat{y}_1|, |y_2 - \hat{y}_2|, |y_3 - \hat{y}_3|, \ldots, |y_n - \hat{y}_n|) \quad (11)$$

*R2 score* determines how well the model would perform in predicting the response variable as shown in (12) where $\bar{y}_l$ denotes the mean value of all predictions. This value is also known as the coefficients of determination. The best possible value is 1 for this case, and the closer to 1, the better model prediction is.

$$R2score = 1 - \frac{\sum_{i=1}^{n}(y_i - \hat{y}_l)^2}{\sum_{i=1}^{n}(y_i - \bar{y}_l)^2} \quad (12)$$

Further feature fit is tested using the residual plot by graphing the residual (the difference between prediction and actual value) vs fitted instance.

## IV. SIMULATIONS AND EXPERIMENTS

### A. Dataset

We collected data from the National Renewable Energy Laboratory (NREL) database available online [23]. The dataset considered for this research contains samples of the three-months-long period starting from May 1, 2018, to July 31, 2018. The raw data entails samples of each minute. It was converted to average hourly instances. Primarily, the dataset had 18 features, among which wind speed in 80m height is our response variable, and other 17 are predictors- solar radiation [listed as global PSP (Precision Spectral Pyranometer)], temperature (2m), estimated sea-level pressure, average wind speed (2m), average wind direction (2m), average wind shear, turbulence intensity, friction velocity, wind chill temperature, dew point temperature, relative humidity, specific humidity, station pressure, average wind speed (5m), accumulated precipitation, atmospheric electric field, and estimated surface roughness. Instances inside "( )" represents the height where the parameter was measured, 'm' stands for meters.

### B. Train-Test Split

The prediction algorithms are trained using a certain dataset. However, the performance of a model depends on how well it can predict the response variable when encounters unknown predictors. Therefore, the dataset is usually divided into two sets: training and test sets. The training dataset is then used to train the prediction algorithm while the test set is allocated to use them as an unknown predictor to analyze the model performance. The ratio of allocating data for training and test is randomly selected, but literature shows 70~80% for training, and 20~30% for the test is common practice [24-26]. In this research, we have separated 80% of the total data to train the models and rest 20% to test the model performance.

### C. Simulation Results

We will discuss the simulation and performances of the state-of-the-art prediction algorithms for wind speed prediction in 80m height for the NREL dataset. We listed the algorithms as Model-1 to 12 in Table 1 and fitted them on the training data for learning. Once the training finished, we evaluated model performances according to the accuracy measures described in Section III on test data.

Normality tests are applied to investigate if the dataset is well modeled (likelihood of data to be normally distributed). In this research, we use the graphical test. In this method, the Chi-Square Quantile-Quantile (Q-Q) plot of multivariate distribution is analyzed to see if the features are normally distributed [27]. If normal, the plot should follow the 45-degree baseline. If not, then normalization is required before fitting the data to any model. We have graphed the Q-Q plot (actual value vs. predicted value of wind speed at 80m) for all models.

For ridge regression, alpha was considered 15 after a few trial and errors. Similarly, for Lasso, the alpha parameter was set to 0.1. For SVR, the default kernel initializer was applied. Table 1 depicts the accuracy measures for each algorithm. Overall, the considered algorithms were able to predict the average wind speed properly with an R2 value greater than 0.9 in most cases, as shown. Among the machine learning algorithms, MAE, MSE, and MedAE are minimum for bagging and random forest regression. Both algorithms show greater accuracy (>96%). Figure 4 illustrates Q-Q plots and respective residual plots (below each Q-Q plot for the same model) for Model 1-9. There is a clear linear pattern in Q-Q plots (for all machine learning algorithms). That verifies the accuracy measures from Table 1, while again bagging and random forest regressions show fewer outliers. The residual plots, in contrast, do not show a linear pattern for any of the models, that supports their accuracy status from Table 1 and validates the feature selection [28]. On the other hand, Models 3-5 show the lowest accuracy among the machine learning regression algorithms with an R2 Score $\approx 0.92$.

Deep learning models- DNN, CNN, and LSTM, are denoted as Model 10-12 in Table 1. Both DNN and CNN use the ReLu activation function. DNN uses thirteen hidden layers, while the neural network of CNN consists of 50 neural network layers. Max pooling size for CNN is 2. On the other hand, LSTM uses a linear activation and consists of 50 hidden layers. In terms of accuracy and error parameters, CNN showed the worst performance, while both DNN and LSTM prediction accuracy were high (>96%). However, LSTM (Model-12) showed the best performance in terms of all metrics; thus, it showed the lowest error terms, while the exact accuracy was 97.8%.

Figure 5 illustrates the deep learning model performances. Prediction visualization, model loss, and MSE are plotted for the models and shown top to down for each model. All three models were run for 500 epochs; however, they reached high accuracy at around 100 epochs. Furthermore, by observing the graphs, it is evident that CNN shows disperse prediction while LSTM is denser. Also, the graphs showing losses and MSE (per epoch) do not show any phenomenon of overfitting or underfitting.

Overall, we can see from Table 1, plots, and discussion, LSTM performed best for our investigation. Therefore, LSTM is the efficient learning algorithm between 12 test models to predict the wind speed at 80m height while the temperature at 2m height, estimated sea-level pressure average wind speed at 2m height, average wind direction at 2m height, average wind shear, turbulence intensity at 2m height, and friction velocity of a certain location are known.

## V. CONCLUSIONS

In this paper, we predicted wind speed at a height that is challenging to reach by using easy to access weather parameters. We investigated twelve artificial intelligence algorithms and concluded that LSTM outperformed other models with 97.8% prediction accuracy. This research will be useful for wind farm planning and feasibility study.

### ACKNOWLEDGMENT

This work was supported by the National Science Foundation (NSF) grants # 1900462 and #1505509, and the DoD grant #W911NF1810475.

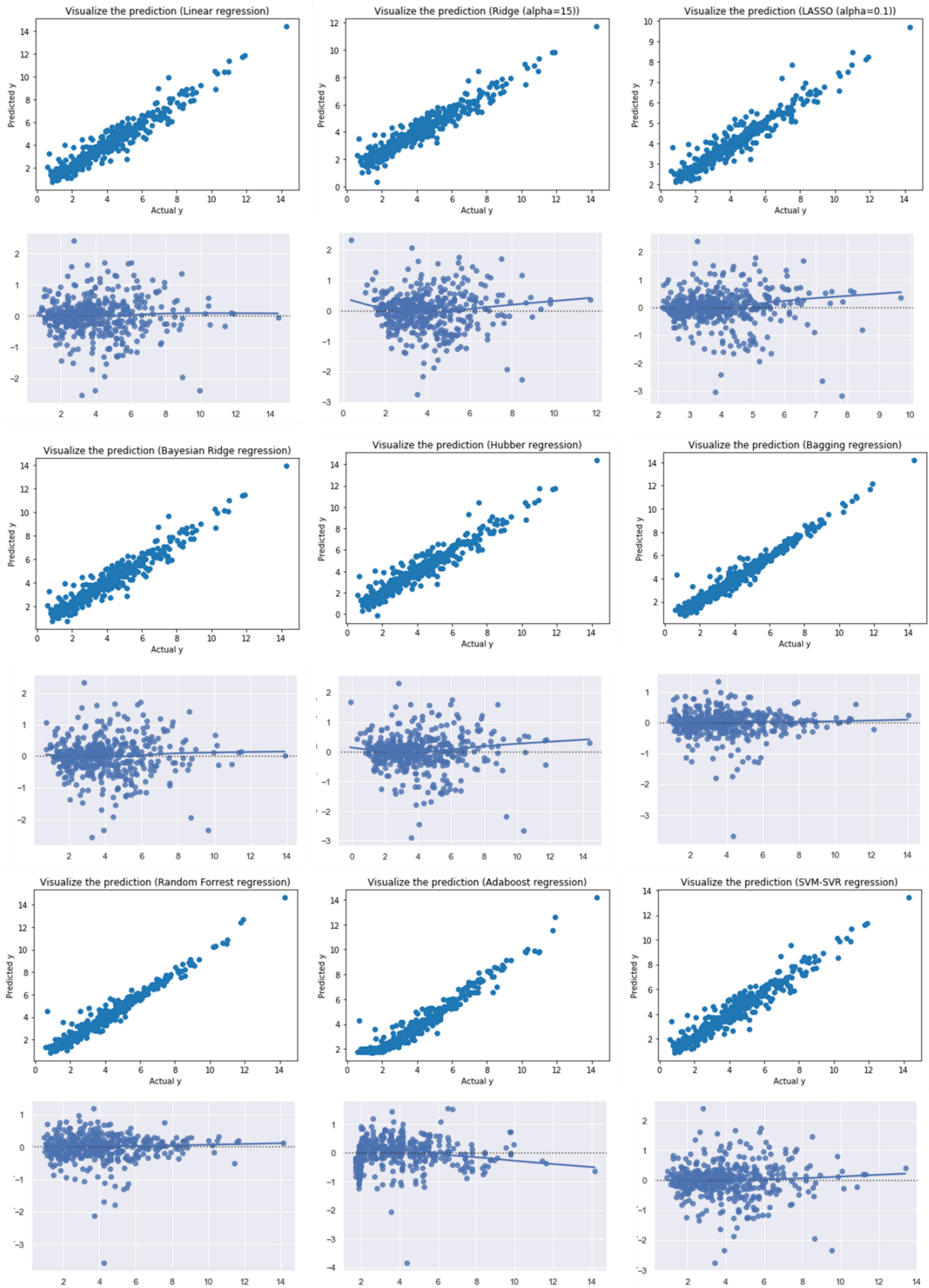

Figure 4 Model 1-9 Prediction Visualization

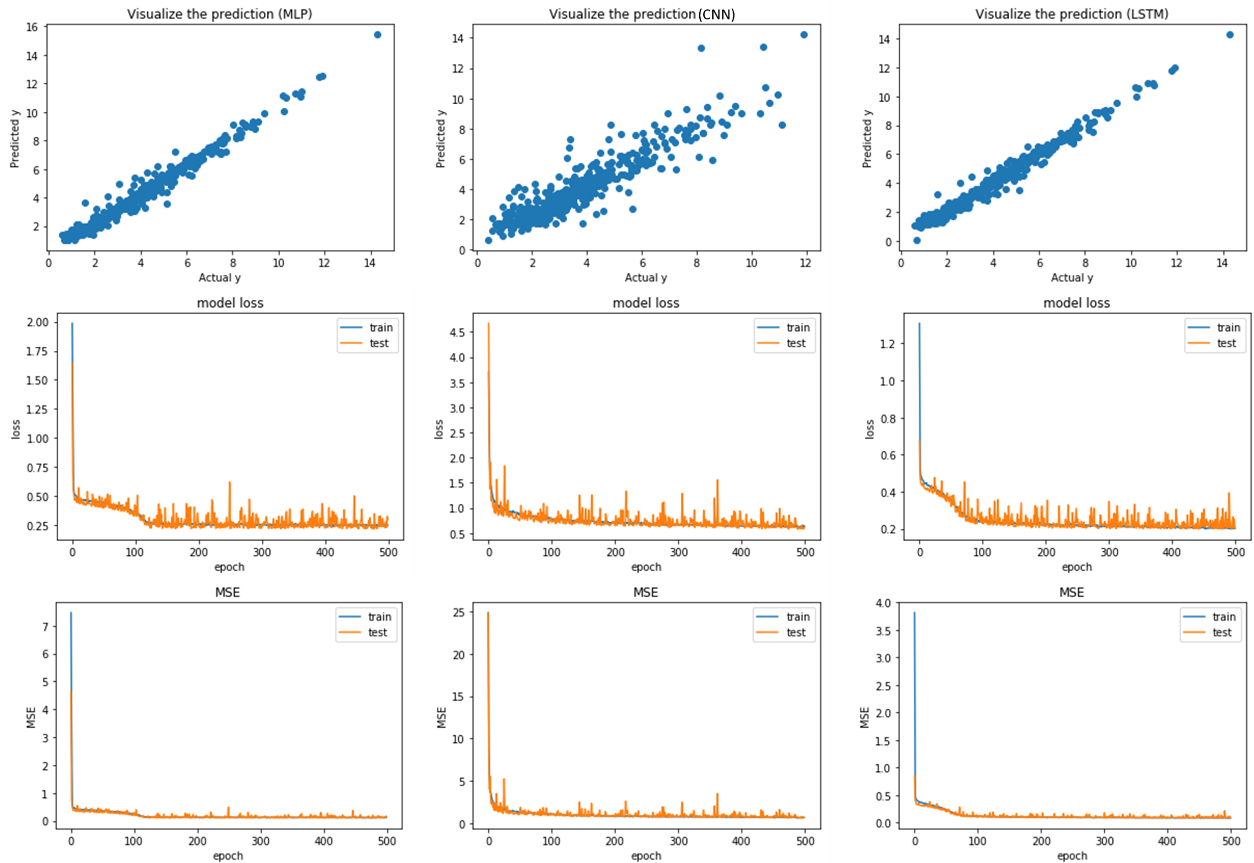

Figure 5 Deep Learning Prediction Visualization

Table 1 Comparative Model Performances

| Model | Algorithm | Mean Absolute Error (MAE) | Mean Squared Error (MSE) | Median Absolute Error (MedAE) | R2 Score |
|---|---|---|---|---|---|
| Model-1 | Multiple linear regression | 0.421 | 0.357 | 0.277 | 0.923 |
| Model-2 | Ridge regression (alpha=0.01) | 0.579 | 0.598 | 0.434 | 0.872 |
| Model-3 | Least absolute shrinkage and selection operator (Lasso) regression (alpha=0.01) | 0.823 | 1.156 | 0.704 | 0.752 |
| Model-4 | Bayesian ridge regression | 0.428 | 0.361 | 0.285 | 0.922 |
| Model-5 | Hubber regression | 0.422 | 0.38 | 0.259 | 0.919 |
| Model-6 | Bagging regression | 0.274 | 0.171 | 0.185 | 0.963 |
| Model-7 | Random forest regression | 0.275 | 0.179 | 0.192 | 0.962 |
| Model-8 | Adaptive boosting (AdaBoost) regression | 0.385 | 0.272 | 0.297 | 0.942 |
| Model-9 | Support vector regression (SVR) | 0.411 | 0.347 | 0.261 | 0.926 |
| Model-10 | Multilayer perceptron (MLP)/ DNN (hidden layer=13, activation=relu) | 0.31 | 0.178 | 0.234 | 0.962 |
| Model-11 | CNN (filters=64, kernel size=2, activation=relu, maxpooling size=2) | 0.634 | 0.831 | 0.45 | 0.82 |
| Model-12 | RNN – LSTM (kernel=normal, activation=linear) | 0.226 | 0.107 | 0.145 | 0.978 |